# Addressing Non-Intervention Challenges via Resilient Robotics utilizing a Digital Twin

Sam T. Harper, Shivoh C. Nandakumar, Daniel Mitchell, Jamie Blanche, Theodore Lim and David Flynn

*Abstract*— Multi-robot systems face challenges in reducing human interventions as they are often deployed in dangerous environments. It is therefore necessary to include a methodology to assess robot failure rates to reduce the requirement for costly human intervention. A solution to this problem includes robots with the ability to work together to ensure mission resilience. To prevent this intervention, robots should be able to work together to ensure mission resilience. However, robotic platforms generally lack built-in interconnectivity with other platforms from different vendors. This work aims to tackle this issue by enabling the functionality through a bidirectional digital twin. The twin enables the human operator to transmit and receive information to and from the multi-robot fleet. This digital twin considers mission resilience and autonomous and human-led decision making to enable the resilience of a multi-robot fleet. This creates the cooperation, corroboration, and collaboration of diverse robots to leverage the capability of robots and support recovery of a failed robot.

## I. Introduction

Intervention offshore is dangerous and costly. If a robot fails offshore, humans are required to intervene by flying out to the robot and recovering it manually. In 2020, there were 11 major injuries in the offshore sector in the UK, with 47 further "over-7-day" injuries reported [1]. Human deployment via chartered helicopter is expensive and has resulted in fatal accidents in the past, reducing these flights has the potential for large cost savings [2]. Mission resilience for remote autonomous operations is an important factor in reducing human interventions in hazardous environments. We define resilience as the ability for an entity to adjust to or recover from misfortune. In terms of autonomous systems, this is the ability of a system to survive such events and still operate under adversity. Through this, mission resilience is achieved as the mission will still succeed. Digital Twin (DT) technologies enable resilience due to increased operational overviews of multi-robot fleet collaboration across resident robots, sensors and environment. This collaboration allows the fleet to meet common mission goals without the need for human intervention.

Currently, to ensure the safe deployment of robotics, humans are required to oversee and recover robots which have failed in hazardous areas, preventing robot loss. A key barrier in the deployment of fully autonomous robots includes autonomous systems which can meet safety compliance requirements. This would allow for imperative requirements where the robot can divert to a safe layby upon detection of a failure. This would allow for a human to safely recover the asset without being exposed to dangerous areas, e.g., confined spaces or nuclear sites. This would reduce necessary interventions, and when these interventions must still occur it ensures safety compliance. Using a Symbiotic Digital Architecture (SDA) will enable safe decisions to unforeseen issues and scenarios to mitigate risk [3].

This research proposes a DT to address non-intervention challenges which currently exist for autonomous platforms in the Offshore Renewable Energy (ORE) sector. This is enabled through automatic recovery of a failed robot. Decision-making within the digital twin can be used to trigger the SDA to preserve mission continuity via adaptive planning. These operations help to ensure mission resilience and minimise the need for human intervention.

To prove the viability of this system, Section I covers the state of the art. In Section II, the methodology of the architecture is explained and an example scenario for a multi-robot fleet operation is presented. This scenario involves a multi-robot asset integrity inspection mission using the Clearpath Husky, DJI Tello and Boston Dynamics Spot platforms. This scenario is then executed within an offshore analogue which represents an offshore substation, results are presented and discussed in Section IV. Finally, there is a conclusion of work done in Section V and future areas of research in Section VI. This work provides an example of the collaboration of off-the-shelf robots, the proof-of-concept video can be accessed at Mitchell et al. [4].

## II. State Of The Art

A limitation in most robotic platforms includes a design for single purpose missions or capabilities; there is not a single robotic platform suited to all tasks of inspection, maintenance, and Repair (IMR). Therefore, it is critical that several different platforms (multi-robot fleet) can communicate and interact with each other whilst sharing results to a DT. However, platforms from different vendors often have different operating systems and requirements. Some platforms, such as Clearpath Husky, use the open-source Robot Operating System (ROS), whereas others, such as Boston Dynamics Spot, have a closed system with a Software Development Kit (SDK) [5] [6]. In cases where the platforms have minimal built-in interconnectivity, they will usually only operate in robotic swarms with other platforms from the same vendor, requiring inter-device software frameworks to communicate [7]. One application of this can be in the form of a DT. DTs date back as far as the 1960s with the Apollo 13 mission and are a well-known concept within the state of the art [8]. Several types of DTs exist for manufacturing, robot, infrastructure and medical [9] [10] [11] [12]. However, there are still challenges that exist in the path of more widespread DT usage. A challenge faced is the lack of standardization across systems. DT solutions currently on the market are bespoke systems made for specific sets of equipment or enterprises. There are no existing, observed, global standards for data formats, data transmission or connectivity concerning DTs. This creates challenges for interoperability – closed and separate systems cannot interact in a symbiotic manner. This poses yet further challenges with respect to digitalization, automation, and integration, which stunts the development of DT technology and methodologies.

Challenges exist with marrying large numbers of assets with the processing power required to monitor the assets. There are solutions capable of processing data from many assets, but these require centralized data processing systems [13], [14], [15], [16]. Conversely, solutions that use low powered hardware are limited in the number of assets they can monitor [17], [18]. Therefore, the flexibility and scalability of hardware for monitoring assets and processing data is a major challenge currently facing DT implementations. Fig. 1 shows a summary of the points raised in this section.

| Knowledge Gap | Opportunity | Description |
|---|---|---|
| Interoperable robotics | Diverse multi-robot fleets | The ability for robotic platforms from different vendors to engage in multi-robot fleet operations |
| Standardisation of Digital Twins | Flexible twins | Enabling digital twins across a wide range of devices and robotic platforms |
| Resilient Robotics | Non-intervention | Reducing or eliminating human intervention in hazardous environments where robots are present |

*Figure 1. List of identified knowledge gaps and opportunities.*

Our proposed Symbiotic System of Systems Approach (SSOSA) seeks to maximize the potential for autonomous mission success via increased resilience, reliability, and safety [3]. Symbiosis is the biological concept of multiple organisms which mutually benefit from their relationships [19]. This stands in opposition to a parasitic relationship, where one organism gains benefit to the detriment of the other. In a symbiotic system, mutualism is achieved through the positive contribution to the health of assets, systems, and robots. This builds further upon safety, resilience, and reliability in dangerous offshore environments.

The SDA is designed to be a scalable, adaptable and platform agnostic architecture, which focusses on bidirectional communications to enable increased transparency in operational decision support. This SDA has been developed primarily aimed at the needs of the ORE sector, but is designed to be flexible for multiple sector-wide use-case scenarios.

The SDA is displayed in Fig. 3 and represents the bidirectional knowledge exchange enabled by the architecture. This allows a human operator to make informed decisions, allowing them to access information about asset integrity inspection and a multi-robot fleet. Conversely, interacting with these processes and platforms through the DT allows the operator to share their knowledge with the robots, through insights gained from the hyper-enabled overview offered by the DT.

The asset integrity inspection block refers to Frequency Modulated Continuous Wave (FMCW) radar scanning of assets. FMCW scanning enables the detection of surface corrosion and subsurface defects within an industrial asset, such as a wind turbine blade or a structural component [20] [21]. This is a useful method of preventative maintenance, as the non-destructive evaluation of these assets leads to early fault detection. Asset integrity inspection is a critical aspect of offshore operations and can be performed autonomously via robotic manipulators.

Mission resilience decision making is defined as the decisions made by the DT to ensure mission resilience. These can be acting upon fault and warning information to ensure another robot can continue the mission, or for performing safety stops to preserve the robots.

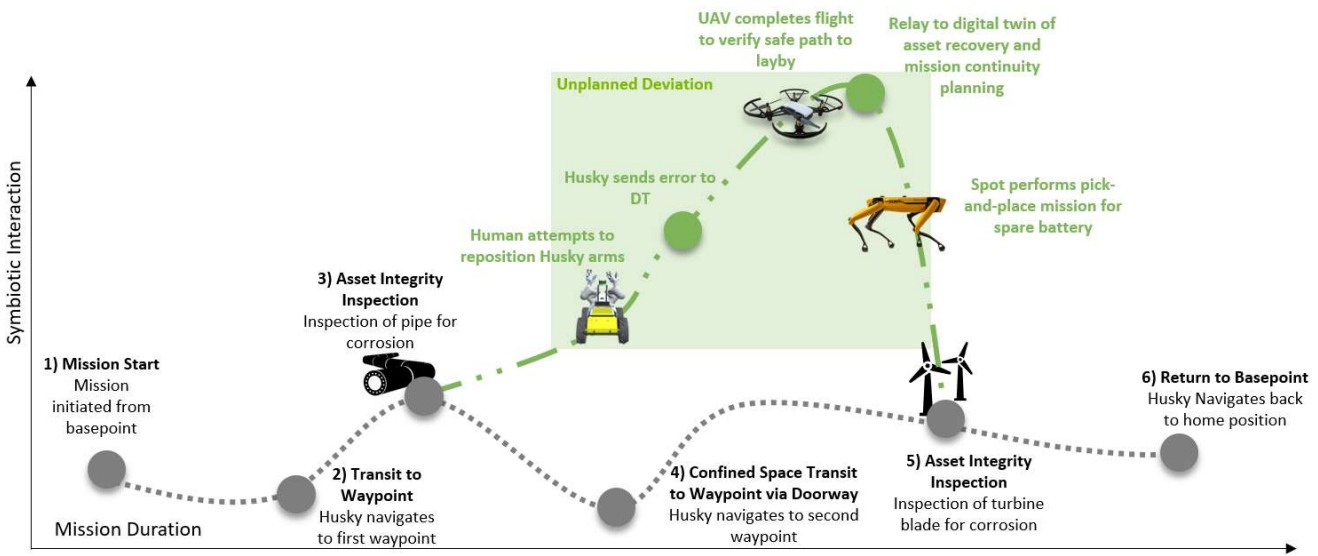

*Figure 2. Graph showing the planned mission and a comparison to the unplanned deviation. The y-axis demonstrates increasing levels of symbiotic interaction between the autonomous platforms and the DT. The deviation highlighted in the box is referenced in Fig. 4.*

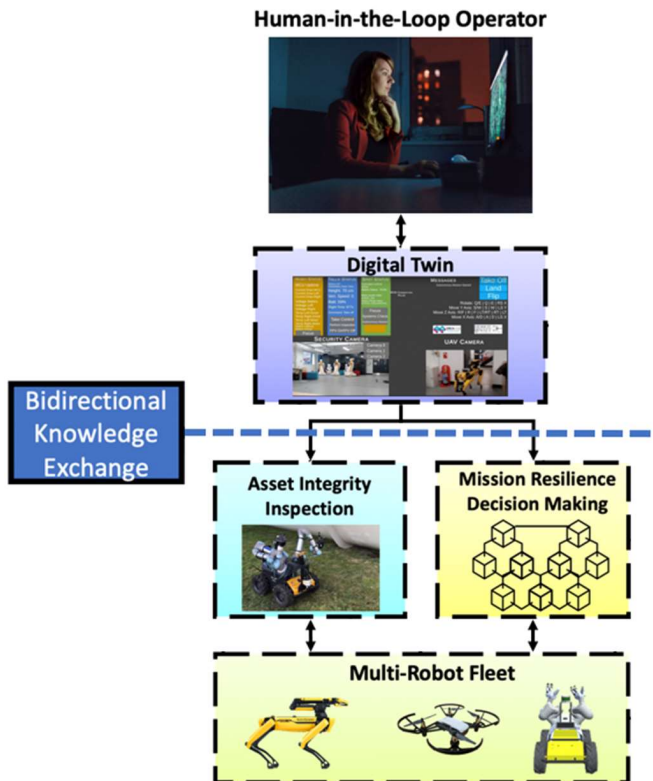

*Figure 3. Graphical representation of the Symbiotic Digital Architecture.*

TABLE I. DEVICES USED IN MISSION

| Device | Description | Image |
|---|---|---|
| Boston Dynamics Spot | Quadruped autonomous platform using a closed-box Python SDK equipped with an arm and six depth cameras. | |
| Clearpath Husky with Dual UR-5 | Wheeled autonomous platform using a ROS core equipped with LIDAR, SLAM, two articulating arms with grippers and a depth camera. | |
| DJI Ryze Tello EDU | Off-the-shelf quadcopter UAV with downwards facing IR sensors for height measurement and a forward-facing camera for video only. | |
| 2020 MacBook Air | Laptop with 8GB RAM and Apple M1 chip. | |

## III. METHODOLOGY

The devices used in this demonstration are shown in Table I. Each robotic platform uses a different operating system and has different capabilities/attributes. This is to demonstrate a multi-robot fleet, where each robotic platform can apply its unique attributes to contribute to mission success via $C^3$ governance (Cooperation, Collaboration and Corroboration).

The DT used in this mission is shown in Fig. 4, named the Operational Decision Support Interface (ODSI). It was developed using the Unity3D development package in C#. Unity enables rapid deployment to multiple devices, ensuring cross-platform compatibility, with alterations only required for the user interface to deal with different input methods. The ODSI acts as a control center for all robotic platforms, giving an immediate overview of the current activities of each platform. This is provided through a view of telemetry from each robot in the fleet and textual updates on mission status. The ODSI contains some decision making which enables symbiotic interactions across the required robotic platforms to complete the mission

The figure is labeled as follows:
1. Status of Husky platform with buttons for pre-determined missions and a focus page.
2. Status of Tello platform with buttons for pre-determined missions and a focus page.
3. Status of Spot platform with buttons for pre-determined missions and a focus page.
4. Messages received from platforms.
5. Security camera footage with camera toggle.
6. Live video feed from Tello platform.
7. Button to trigger the multi-robot corrosion inspection mission (Fig. 6).

The focus page is demonstrated in Fig. 5 and is designed to give a more in-depth overview tailored to each robotic platform. The pictured example with the Husky shows the ability to position the arms with a "ghosting" interface, where the sliders control arm movement and the 3D model displaying the current (ghost) and desired (solid) arm positions. These features give the operator a hyper-enabled overview of the mission status, which is essential in Beyond Visual Line of Sight (BVLOS) operations. Corroboration between the data given by the ODSI and a visual overview of operations increases the operators trust in the system, so are necessary to include in the ODSI.

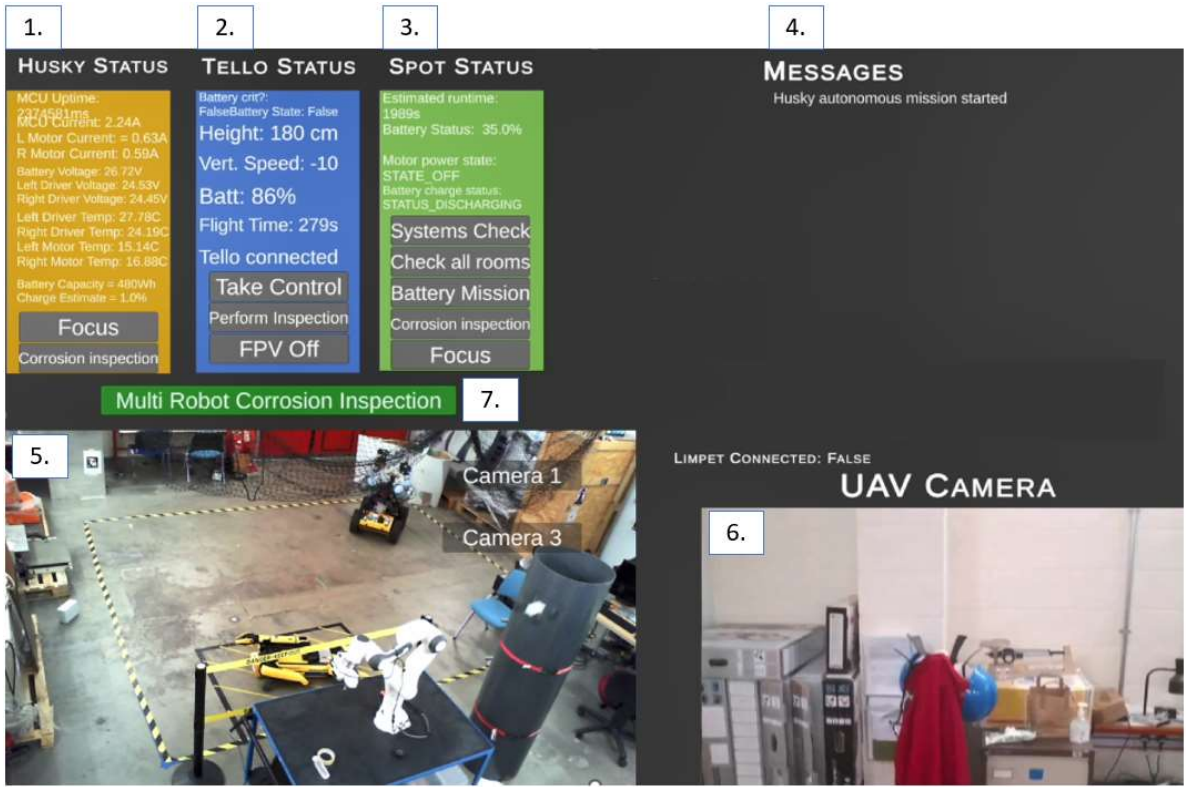

*Figure 4. The ODSI, with numbers superimposed corresponding to features within the DT.*

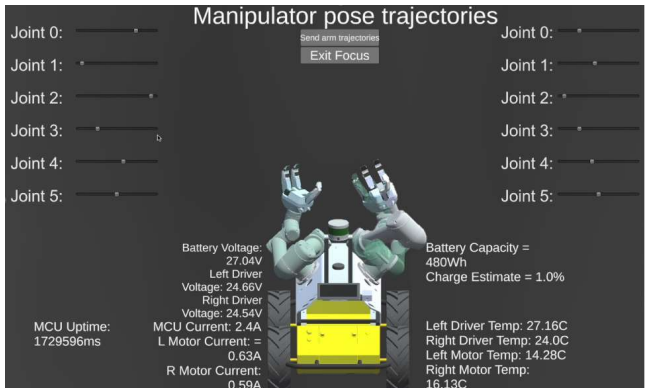

*Figure 5. ODSI "Focus" page example for Husky, with ghosted right arm position*

To demonstrate mission resilience, a scenario mission was planned as shown in Fig. 6. The objective of this mission is to perform two asset integrity inspections of a corroded metal sheet and a wind turbine blade. These were chosen to mirror the types of asset integrity inspections completed in the ORE industry. In the validation and assessment of the resilience of the autonomous mission, the mission resilience decision making element is the focus of this investigation. This is achieved by running the mission twice, once in a "perfect" state where there are no issues, and once again with an induced failure to ensure mission resilience. This fault is induced on the Husky robotic platform as a battery fault when the user requests manual repositioning of the arms. This allows the decision-making element to decide on which predetermined action to take. In the case that a warning is triggered, another robot will move to assist the stranded robot.

The information is fed to a human-in-the-loop operator to draw on their expertise. To allow successful completion of the mission, the mission resilience decision making element alerts Spot to perform a battery replacement mission.

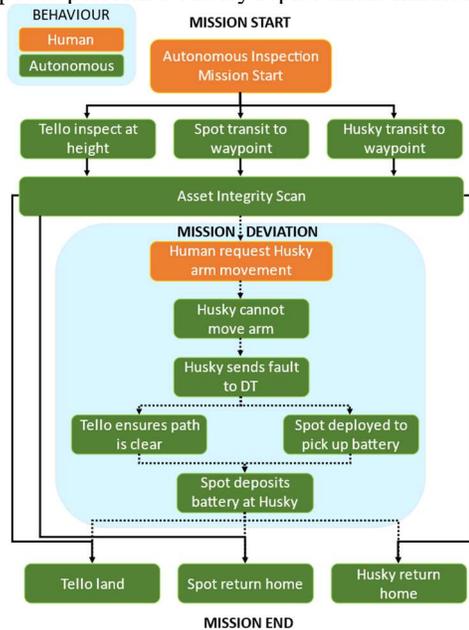

*Figure 6. Resilience mission plan. Area highlighted by blue box refers to the deviation shown in Fig. 2.*

The cyber physical system layout of this mission is displayed in Fig. 7. This layout uses a client-server architecture, with a DT host laptop acting as the server of all communications, using a separate port for each platform. The

Tello platform is limited as it can only use a direct Wi-Fi communication to the DT host laptop, therefore the host must be connected to a router via Ethernet to allow communications with all other platforms. Wi-Fi is used to connect to the Spot and Husky via the router. Cameras connected via USB act as security cameras giving different static viewpoints of the area for improved $C^3$. For the purposes of demonstration, all connections are over a Local Area Network (LAN), future implementation work is to use the system over the internet.

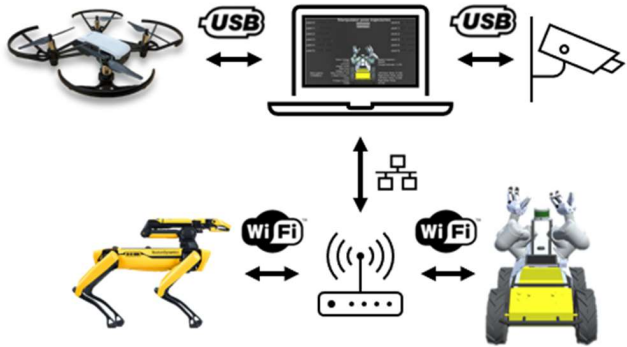

*Figure 7. Graphical representation of network layout of the mission.*

For communication between each platform and the DT, TCP sockets are used. Messages are sent as short strings and decoded on each side as required. An example of this is shown in Fig. 8, where a mission can be triggered from a preset list using a single character string sent from the host DT. It also shows the typical format in which robot information is sent back to the DT, allowing data packets to be kept small in bandwidth constrained scenarios, which are likely when communicating with offshore platforms.

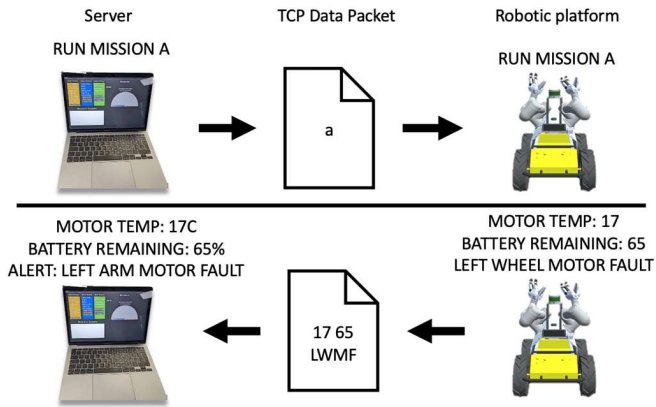

*Figure 8. Graphical representation of two-way communication between the DT and an autonomous platform for a Left Wheel Motor Fault (LWMF).*

## IV. RESULTS & DISCUSSION

This work represents a proof of concept for mitigating the risks that exist in reliability, resilience, and safety of autonomous robotic platforms. To demonstrate this, the missions were run as outlined in Fig. 6. The demonstration was performed in an offshore analogue space, as shown in Fig. 9.

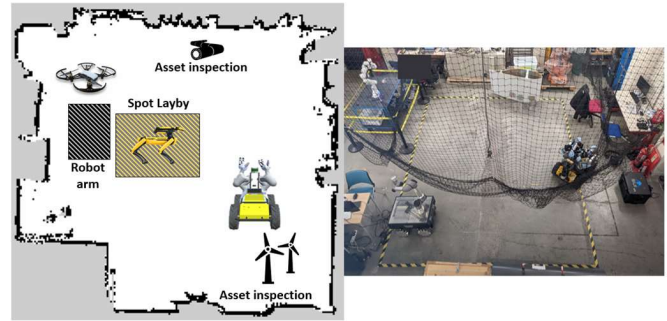

*Figure 9. Left: Diagram of the mission space. Right: Photograph of the mission space used.*

The mission resilience decision making block (Fig. 3) was evaluated via this demonstration, where the actions of the multi-robot fleet were not hard coded in the sequence of events of the mission. The fault was induced on the Husky platform once the command to move the arms was sent, and the Husky informed the DT of this through a message. This information was shared with the digital twin and mission resilience decision making block. At this stage, the deviations of the mission occurred via a pre-determined remedial action. In this case, the Spot performed an automated pick-and-place mission with the spare battery to enable resilience for Husky.

The "perfect" mission run-through showed the use case for an SDA via the cooperation of the robotic platforms to perform the asset integrity mission. This is shown visually in Fig. 10 using images from the security cameras.

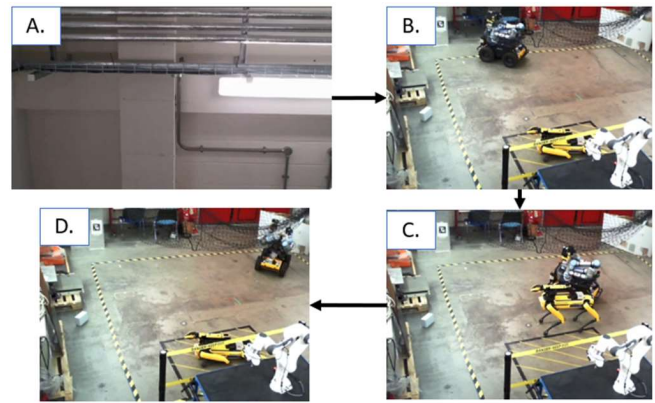

*Figure 10. "Perfect" run through. Demonstrates: A. Tello inspection at height (view is from camera on the front of Tello), B. Husky asset inspection, C. Husky and Spot asset inspection, D. All return to home positions*

The failure demonstration proved the use of an SDA for DT. Corroboration was achieved in the Tello to establish a safe path to the layby. Collaboration was achieved through Spot providing the battery analogue to Husky. Cooperation was achieved through all robotic platforms sharing information to the DT. The reactions of the DT to the simulated robot failures enabled other robots to carry out tasks and ensure the resilience of the failed robot. In a real environment, this would enable the safe recovery of the robot from a hazardous environment. The mission continuity afforded by the cooperation, corroboration and collaboration of the other platforms allowed the recovery mission to be carried out successfully without on-site human intervention. This is shown visually in Fig. 11.

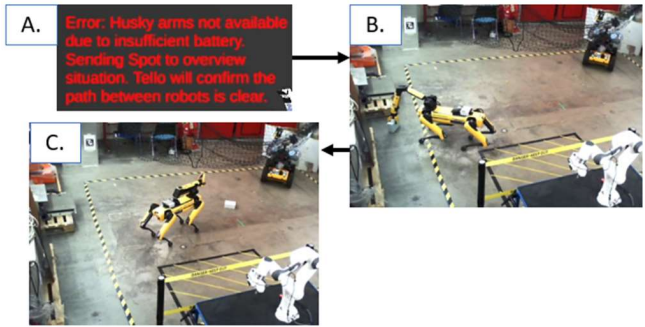

*Figure 11. Failure run through. Demonstrates: A. Failure message displayed on ODSI, B. Spot picking up battery box, C. Spot depositing battery box at Husky.*

## V. Conclusion

This work has presented a DT which enables bidirectional communications of a multi-robot fleet, increasing robotic resilience. This has been achieved via an SDA, which includes a diverse range of robots and infrastructural sensors. The advantages of utilizing a multi-robot fleet include the leveraging of robotic capabilities to achieve a mutual goal: safe recovery of a robot, representing an advance in autonomous collaboration.

Of key importance is to ensure that multi-robot fleets do not become an operational risk for human intervention and recovery of assets. The development of fault detection and monitoring via a runtime reliability ontology represents an opportunity to address these non-intervention challenges, which inhibit BVLOS missions. Symbiotic digitalization enables for improved human-machine interactions across remote robotic agents. This leads to improved control and awareness for a human operator via increased resilience, reliability, and safety.

## VI. Future Work

There is scope to add more complex Planning Domain Definition Language (PDDL) based planning to the SDA. Implementation work on this has been carried out by Carreno *et al.* on their work in extreme environments [22]. This would improve the dynamic robot planning, allowing for consideration of temporal aspects of planning and include situational awareness autonomously. The work is currently limited to simulation and has not been applied via implementation to physical service robots.

The manipulation of objects by robots needs to be improved to allow for true non-intervention and resilience. Currently, it is possible to use robotic manipulators as tools to pick and place objects, however it is not currently feasible for a mobile robot to perform maintenance operations on another mobile robot e.g., intervening internally on a robot and performing a battery replacement. Development in this field is critical for resident robotic operations BVLOS.

Using the system over the internet as a Wide Area Network (WAN) instead of a LAN will provide more opportunity for BVLOS operations, due to the possibility of fully remote operations, and will require full control over the network firewall and port forwarding settings. Such control strategies were outside the scope of this demonstration, as the focus was on displaying the benefits of a symbiotic multi-robot fleet.

Trust is also an issue relating to digital twins – how can an operator trust a BVLOS system without seeing it first-hand? It is vital that any simulations in a DT are based on realistic and correct models [23]. This helps to ensure that a DT will fully correlate to what is happening in the physical space. As such, trust is essential for decision making support, so standardization and progress in this area is essential for DT development [24].

## VII. Acknowledgments

The research within this paper has been supported by the Offshore Robotics for Certification of Assets (ORCA) Hub [EP/R026173/1] and the authors want to thank MicroSense Technologies Ltd for use of their patented FMCW radar sensor (PCT/GB2017/053275).

## VIII. References

[1] Energy Division, Health and Safety Executive, "Annual offshore statistics and regulatory activity report 2020," Health and Safety Executive, 2020.

[2] F. Menezes, O. Porto, M. L. Reis, L. Moreno, M. P. d. Aragão, E. Uchoa, H. Abeledo and N. C. d. Nascimento, "Optimizing Helicopter Transport of Oil Rig Crews at Petrobras," *Interfaces*, vol. 40, no. 5, pp. 408-416, 2010.

[3] D. Mitchell, O. Zaki, J. Blanche, J. Roe, L. Kong, S. Harper, V. Robu, T. Lim and D. Flynn, "Symbiotic System of Systems Design for Safe and Resilient Autonomous Robotics in Offshore Wind Farms," 02 04 2021. [Online]. Available: https://arxiv.org/abs/2101.09491. [Accessed 17 06 2021].

[4] Smart Systems Group, "A Scalable Cyber Physical Architecture for Symbiotic Multi-Robot Fleet Autonomy - YouTube," Smart Systems Group, 1 July 2022. [Online]. Available: https://youtu.be/ktPZb9ZC2Q0. [Accessed 14 September 2022].

[5] ROS.org, "Robots/Husky - ROS Wiki," 15 July 2019. [Online]. Available: http://wiki.ros.org/Robots/Husky. [Accessed 14 February 2022].

[6] Boston Dynamics, "Spot SDK — Spot 3.0.3 documentation," 2022. [Online]. Available: https://dev.bostondynamics.com/. [Accessed 14 February 2022].

[7] H. Hong, W. Kang and S. Ha, "Software Development Framework for Cooperating Robots with High-level Mission Specification," in *2020 IEEE/RSJ*

*International Conference on Intelligent Robots and Systems (IROS)*, Las Vegas, 2020.

[8] Siemens, "Apollo 13: The First Digital Twin | Simcenter," Siemens, 14 04 2020. [Online]. Available: https://blogs.sw.siemens.com/simcenter/apollo-13-the-first-digital-twin/. [Accessed 18 05 2020].

[9] T. Mori, K. Ikeda, N. Takeshita, K. Teramura and M. Ito, "Validation of a novel virtual reality simulation system with the focus on training for surgical dissection during laparoscopic sigmoid colectomy," *BMC Surgery,* vol. 22, no. 1, 2022.

[10] È. Pairet, P. Ardón, X. Liu, J. Lopes, H. Hastie and K. S. Lohan, "A Digital Twin for Human-Robot Interaction," in *2019 14th ACM/IEEE International Conference on Human-Robot Interaction (HRI)*, Daegu, South Korea, 2019.

[11] M. Shahzad, M. T. Shafiq, D. Douglas and M. Kassem, "Digital Twins in Built Environments: An Investigation of the Characteristics, Applications, and Challenges," *Building,* vol. 12, no. 2, p. 120, 2022.

[12] Y. Zhang, C. Zhang, J. Yan, C. Yang and Z. Liu, "Rapid construction method of equipment model for discrete manufacturing digital twin workshop system," *Robotics and Computer-Integrated Manufacturing,* vol. 75, p. 102309, 2022.

[13] K. T. Park, Y. W. Nam, H. S. Lee, S. J. Im, S. D. Noh, J. Y. Son and H. Kim, "Design and implementation of a digital twin application for a connected micro smart factory," *International Journal of Computer Integrated Manufacturing,* vol. 32, no. 6, pp. 596-614, 2019.

[14] L. Hernández and S. Hernández, "Application of digital 3D models on urban planning and highway design," *Transactions on the Built Environment,* vol. 30, pp. 391-402, 1997.

[15] S. Sierlaa, V. Kyrkia, P. Aarnioa and V. Vyatkinab, "Automatic assembly planning based on digital product descriptions," *Computers in Industry,* vol. 97, pp. 34-46, 2018.

[16] M. Armendia, A. Alzaga, F. Peysson, T. Fuertjes, F. Cugnon, E. Ozturk and D. Flum, "Machine Tool: From the Digital Twin to the Cyber-Physical Systems," in *Twin-Control: A Digital Twin Approach to Improve Machine Tools Lifecycle*, Springer, 2019, pp. 3-21.

[17] S. Rabah, A. Assila, E. Khouri, F. Maier, F. Ababsa, V. Bourny, P. Maier and F. Merienne, "Towards improving the future of manufacturing through digital twin and augmented reality technologies," in *28th International Conference on Flexible Automation and Intelligent Manufacturing*, Columbus, Ohio, 2018.

[18] A. d. S. Barbosa, F. P. Silva, L. R. d. S. Crestani and R. B. Otto, "Virtual Assistant to Real Time Training on Industrial Environment," *Advances in Transdisciplinary Engineering,* vol. 7, pp. 33-42, 2018.

[19] A. McConnell, D. Mitchell, K. Donaldson, S. Harper, J. Blanche, T. Lim, D. Flynn, A. Stokes, S. Tushar and F. Iqbal, "The Future Workplace: A Symbiotic System of Systems Environment," in *Cyber-Physical Systems Solutions to Pandemic Challenges*, Boca Raton, CRC Press, 2022, pp. 254-325.

[20] D. Mitchell, J. Blanche and D. Flynn, "An Evaluation of Millimeter-wave Radar Sensing for Civil Infrastructure," in *2020 11th IEEE Annual Information Technology, Electronics and Mobile Communication Conference (IEMCON)*, Vancouver, 2020.

[21] J. Blanche, D. Mitchell, R. Gupta, A. Tang and D. Flynn, "Asset Integrity Monitoring of Wind Turbine Blades with Non-Destructive Radar Sensing," in *2020 11th IEEE Annual Information Technology, Electronics and Mobile Communication Conference (IEMCON)*, Vancouver, 2020.

[22] Y. Carreno, J. Scharff Willners, Y. Petillot and R. P. Petrick, "Situation-Aware Task Planning for Robust AUV Exploration in Extreme Environments," in *Proceedings of the IJCAI Workshop on Robust and Reliable Autonomy in the Wild*, Montreal, 2021.

[23] K. Wärmefjord, R. Söderberg, B. Schleich and H. Wang, "Digital twin for variation management: A general framework and identification of industrial challenges related to the implementation," *Applied Sciences (Switzerland),* vol. 10, no. 10, p. 3342, 2020.

[24] S. Atkinson, "Collaborative trust networks in engineering design adaptation," in *ICED 11 - 18th International Conference on Engineering Design - Impacting Society Through Engineering Design*, Copenhagen, Denmark, 2011.